\documentclass[10pt,twocolumn,letterpaper]{article}

\usepackage[pagenumbers]{cvpr} 

\definecolor{cvprblue}{rgb}{0.21,0.49,0.74}
\usepackage[pagebackref,breaklinks,colorlinks,allcolors=cvprblue]{hyperref}

\def\paperID{} 
\def\confName{CVPR}
\def\confYear{2026}

\title{UAV-Assisted Scan-to-Simulation for Landslides Using Physics-Informed Gaussian Splatting}

\author{Zhenyu Liang\\
HKUST\\
\and
Jack C.P. Cheng\\
HKUST\\
}

\begin{document}
\maketitle
\begin{abstract}
Landslide monitoring and simulation play an important role in urban safety assessment and disaster prevention. Existing landslide simulation pipelines typically rely on digital elevation model and mesh-based representations, which are suitable for geometric analysis, but often lack visual realism. This limitation reduces their effectiveness in interactive applications, hazard communication, and public education. In this paper, we propose a UAV-based scan-to-simulation framework that bridges photorealistic scene capture and physics-based landslide simulation through 3DGS. Specifically, our pipeline includes four stages: (1) UAV-based acquisition of slope imagery, (2) reconstruction of a low-anisotropy 3DGS scene representation, (3) volumetric conversion of the target simulation region by filling the interior of the surface-based model, and (4) integration with the Material Point Method (MPM) for landslide simulation. We validate the proposed framework on a real landslide site in Hong Kong that experienced a severe landslide event. The results show that our method supports both realistic visual reconstruction and effective simulation.
\end{abstract}    
\section{Introduction}
\label{sec:intro}

Robotic sensing of real-world scenes and infrastructure has become an important tool in civil engineering for safety assessment, compliance inspection, and disaster prevention. Recent advances in UAV-based image reconstruction and point-cloud scanning have made large-scale site acquisition increasingly efficient and accessible. However, most existing pipelines use scanned data primarily for texture-based inspection, geometry-based checking, or conversion to object-level information models \cite{lv2026automating}. As a result, scanned scenes are often treated as documentation assets rather than simulation-ready representations.

However, in many civil engineering applications, appearance and geometry alone are insufficient. The safety and failure of infrastructure are ultimately governed by mechanics, which requires physics-based analysis beyond visual inspection or geometric verification. This limitation is particularly critical in slope environments, where landslides pose significant threats to human life and built assets. Traditional landslide simulation methods typically transform scanned digital elevation model data into mesh-based models for numerical analysis \cite{zhao2023multiscale}. While effective for engineering purposes, such representations often lack visual realism and are therefore difficult for non-experts to interpret.


Recent progress in Gaussian Splatting \cite{kerbl20233d} offers a promising foundation toward this goal. By representing scenes as collections of Gaussian ellipsoids, 3D Gaussian Splatting (3DGS) enables photorealistic modeling and real-time rendering, while also providing a particle-like representation that is naturally compatible with simulation paradigms for granular or deformable media. However, the original 3DGS formulation is designed primarily for static scene rendering and does not directly support physically consistent dynamic simulation.

\begin{figure}[t]
  \centering
   \includegraphics[width=0.8\linewidth]{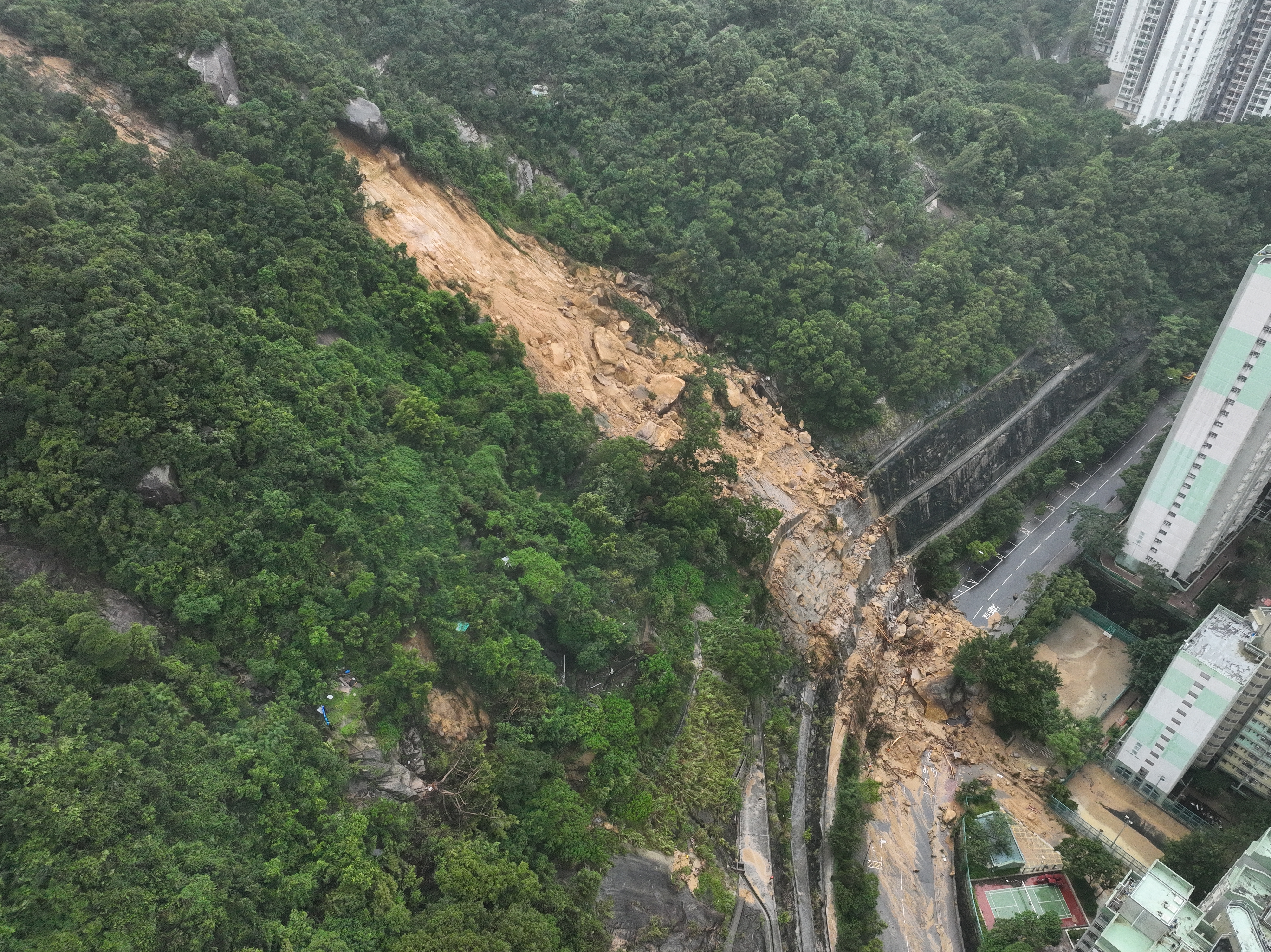}

   \caption{Severe landslide occurred at Yiu Hing Road, Shau Kei Wan, Hong Kong.}
   \label{fig:SKW}
\end{figure}

To address this gap, we propose a UAV-based scan-to-simulation framework for landslides using physics-informed Gaussian Splatting. Our framework includes four stages: UAV-based slope data acquisition, low-anisotropy 3DGS reconstruction, volumetric conversion of the target simulation region via interior filling, and physics-based simulation through integration with the Material Point Method (MPM). We validate the framework at a real landslide site on Yiu Hing Road, Shau Kei Wan, Hong Kong, where a severe landslide occurred during the Black Rainstorm on 8 September 2023, as shown in Figure \ref{fig:SKW}. The results demonstrate the effectiveness of our method for combining photorealistic scene reconstruction with physically grounded landslide simulation.

\section{Preliminaries}
\label{sec:preliminaries}

We briefly review the two foundations of our framework: 3DGS for photorealistic scene representation and rendering, and the MPM for physics-based simulation of landslide dynamics.

\subsection{3D Gaussian Splatting}
\label{sec:3DGS}

3D Gaussian Splatting (3DGS) \cite{kerbl20233d} represents a scene as a set of anisotropic Gaussian primitives, where each Gaussian is parameterized by a center $\boldsymbol{\mu}$, covariance $\boldsymbol{\Sigma}$, opacity $\alpha$, and view-dependent color. The spatial density of a Gaussian is defined as

\begin{equation}
G(\mathbf{x}) = \exp \left( -\frac{1}{2} (\mathbf{x} - \boldsymbol{\mu})^\top \boldsymbol{\Sigma}^{-1} (\mathbf{x} - \boldsymbol{\mu}) \right),
\end{equation}
where $\mathbf{x} \in \mathbb{R}^3$. The covariance is commonly decomposed as

\begin{equation}
\boldsymbol{\Sigma} = \mathbf{R}\mathbf{S}\mathbf{S}^\top\mathbf{R}^\top,
\end{equation}
with rotation matrix $\mathbf{R}$ and diagonal scaling matrix $\mathbf{S}$.

For rendering, each 3D Gaussian is projected onto the image plane as a 2D Gaussian. Let $\mathbf{J}$ denote the Jacobian of the projection function and $\mathbf{W}$ the world-to-camera transform. The projected covariance is approximated by

\begin{equation}
\boldsymbol{\Sigma}' = \mathbf{J}\mathbf{W}\boldsymbol{\Sigma}\mathbf{W}^\top\mathbf{J}^\top.
\end{equation}

Pixel colors are then obtained by alpha compositing along each ray:

\begin{equation}
C = \sum_{i=1}^{N} T_i \alpha_i c_i, \qquad
T_i = \prod_{j=1}^{i-1}(1-\alpha_j),
\end{equation}
where $c_i$ is the color of the $i$-th Gaussian and $T_i$ is the accumulated transmittance. The Gaussian parameters are optimized from multi-view images using an image reconstruction loss $\mathcal{L}_{\mathrm{rec}}$, typically combining $\mathcal{L}_1$ and SSIM terms.

\subsection{Material Point Method}
\label{sec:MPM}

The Material Point Method (MPM) \cite{zhao2023multiscale} is a hybrid Eulerian--Lagrangian method for simulating continuum materials with large deformation. It discretizes the material into Lagrangian particles carrying physical states, while computing forces and momentum updates on a background Eulerian grid. This design makes MPM particularly suitable for landslide simulation.

Let $p$ index particles and $i$ index grid nodes. Each particle carries mass $m_p$, position $\mathbf{x}_p$, velocity $\mathbf{v}_p$, and deformation gradient $\mathbf{F}_p$. In each step, particle mass and momentum are transferred to the grid through interpolation weights $w_{ip}$:

\begin{equation}
m_i = \sum_p w_{ip} m_p, \qquad
m_i \mathbf{v}_i = \sum_p w_{ip} m_p \mathbf{v}_p.
\end{equation}

The internal force on grid node $i$ is computed as

\begin{equation}
\mathbf{f}^{\mathrm{int}}_i = -\sum_p V_p \boldsymbol{\sigma}_p \nabla w_{ip},
\end{equation}
where $V_p$ and $\boldsymbol{\sigma}_p$ denote the particle volume and Cauchy stress, respectively. With external force $\mathbf{f}^{\mathrm{ext}}_i$, the grid velocity is updated by

\begin{equation}
\mathbf{v}_i^{n+1} = \mathbf{v}_i^n + \Delta t \frac{\mathbf{f}^{\mathrm{int}}_i + \mathbf{f}^{\mathrm{ext}}_i}{m_i}.
\end{equation}

The updated grid velocity is then transferred back to particles:

\begin{equation}
\mathbf{v}_p^{n+1} = \sum_i w_{ip} \mathbf{v}_i^{n+1}, \qquad
\mathbf{x}_p^{n+1} = \mathbf{x}_p^n + \Delta t \sum_i w_{ip} \mathbf{v}_i^{n+1}.
\end{equation}

The deformation gradient is updated as

\begin{equation}
\mathbf{F}_p^{n+1} =
\left(
\mathbf{I} + \Delta t \sum_i \mathbf{v}_i^{n+1} (\nabla w_{ip})^\top
\right)\mathbf{F}_p^n.
\end{equation}

By combining particle-based material tracking with grid-based computation, MPM naturally handles large deformation, separation, and flow-like behavior, making it a suitable physics engine for landslide simulation in our framework.

\begin{figure*}[t]
  \centering
  \includegraphics[width=\textwidth]{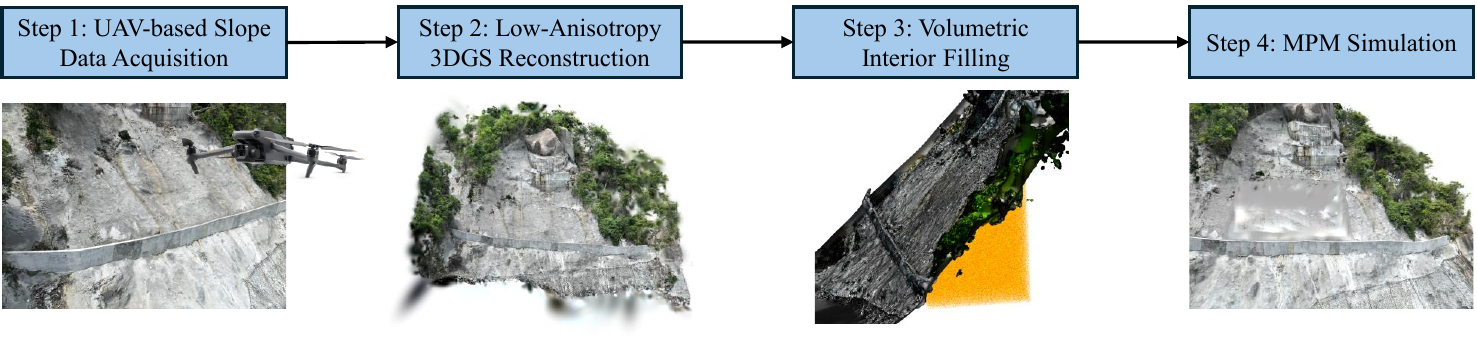}
  \caption{The proposed UAV-based scan-to-simulation framework for landslides using physics-informed Gaussian Splatting.}
  \label{fig:pipeline}
\end{figure*}
\section{Methodology}
\label{sec:methodology}

This section introduces our proposed scan-to-simulation pipeline for landslides from UAV observations using physics-informed Gaussian splatting. The overall framework is illustrated in Figure~\ref{fig:pipeline}.

\subsection{UAV-based Slope Data Acquisition}
\label{subsec:data_acquisition}

We acquire slope data using a DJI M3E drone, whose RTK module provides centimeter-level positioning accuracy. To ensure sufficient coverage for 3D reconstruction, image capture follows two principles. First, the slope is photographed at multiple distances, including both far-range and close-range views, so that geometry at different spatial scales can be reliably observed. Second, the UAV captures images from multiple viewing angles around the slope to achieve dense view coverage and reduce occlusions. These acquisition strategies provide the multi-scale and multi-view observations required for robust 3DGS reconstruction.

To further improve camera pose accuracy, we explicitly utilize the GPS metadata recorded by the UAV. Specifically, each image is associated with pose information in the WGS84 coordinate system, including latitude, longitude, altitude, and pitch-yaw-roll angles. We convert these geo-referenced poses into a local East-North-Up coordinate system and then transform them into a COLMAP-compatible camera format. In this way, the UAV's high-precision localization capability is incorporated into the reconstruction pipeline.

\begin{figure*}[t]
  \centering
  \includegraphics[width=\textwidth]{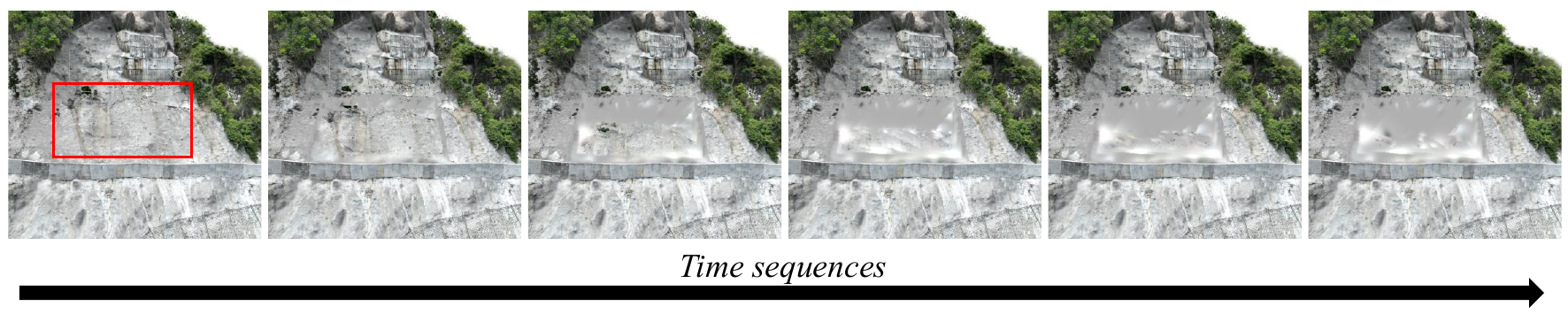}
  \caption{Qualitative performance of landslide simulation using the proposed framework.}
  \label{fig:result}
\end{figure*}

\subsection{Low-Anisotropy 3DGS Reconstruction}
\label{subsec:low_aniso_3dgs}

Given the captured images and the corresponding camera poses in COLMAP-format, we reconstruct the slope scene using 3DGS. However, unlike standard novel-view synthesis settings, our reconstructed Gaussians undergo motion and rotation during physical simulation. In this case, highly anisotropic Gaussian ellipsoids may produce undesirable rendering artifacts such as elongated streaks or spiky structures when they translate or flip during simulation. Therefore, in addition to the reconstruction objective introduced in Section~\ref{sec:3DGS}, we impose an anisotropy regularization term to constrain the aspect ratios of Gaussian ellipsoids.

Let $S_p$ denote the three principal scales of the $p$-th Gaussian, and let $r$ denote a prescribed upper bound on the allowed scale ratio. We define the anisotropy loss as

\begin{equation}
\mathcal{L}_{\mathrm{aniso}}
=
\frac{1}{|\mathcal{P}|}
\sum_{p \in \mathcal{P}}
\left(
\max \left\{
\frac{\max(S_p)}{\min(S_p)},\; r
\right\}
- r
\right),
\end{equation}
where $\mathcal{P}$ denotes the set of Gaussians. This loss penalizes Gaussians whose scale ratio exceeds the threshold $r$, while leaving near-isotropic Gaussians unaffected. In our implementation, we set the maximum ratio to $r = 3$.

The final reconstruction objective is written as

\begin{equation}
\mathcal{L}
=
\mathcal{L}_{\mathrm{rec}}
+
\lambda_{\mathrm{aniso}} \mathcal{L}_{\mathrm{aniso}},
\end{equation}
where $\lambda_{\mathrm{aniso}}$ controls the strength of anisotropy regularization. 

\subsection{Volumetric Interior Filling}
\label{subsec:volumetric_filling}

The reconstructed 3DGS model from image observations mainly represents the visible surface of the slope. However, landslide simulation requires a volumetric interior rather than a hollow surface shell. Therefore, we introduce a volumetric interior filling step to synthesize subsurface Gaussian primitives beneath the reconstructed slope surface.

We first treat the Gaussian centers of the reconstructed 3DGS model as a point cloud and extract a surface mesh using Poisson reconstruction. Based on the target simulation region, we define a filling domain bounded by $(x_{\min}, x_{\max}, y_{\min}, y_{\max}, z_{\min})$, while using the reconstructed mesh surface as the top boundary in the vertical direction. Within this domain, we uniformly distribute a set of interior Gaussian centers to populate the soil volume below the observed surface.

For each interior Gaussian center, we determine its size according to the distance to its nearest point on the reconstructed surface. Specifically, let $d_p$ denote the nearest-surface distance for the $p$-th interior Gaussian. We set its short axis, middle axis, and long axis as

\begin{equation}
s_p^{\mathrm{short}} = d_p, \qquad
s_p^{\mathrm{mid}} = 1.5 d_p, \qquad
s_p^{\mathrm{long}} = 2 d_p.
\end{equation}

 For appearance, we assign each interior Gaussian a color consistent with the slope surface. To avoid view-dependent color inconsistency during motion and rotation, we only use the base color term of spherical harmonics (SH) and discard higher-order SH coefficients for the filled interior Gaussians.

\subsection{MPM Simulation}
\label{subsec:mpm_simulation}

After obtaining the volumetric Gaussian model, we perform landslide simulation using the Material Point Method. Specifically, we assign material properties, external forces such as gravity, and boundary conditions to the filled slope volume, and then simulate its dynamic evolution according to the MPM formulation introduced in Section~\ref{sec:MPM}. In our implementation, we adopt the MPM solver provided in~\cite{xie2024physgaussian}.

Given the initialized particles and physical parameters, the MPM solver computes particle motion over time under the specified forces and constraints. This produces the temporal evolution of the landslide process, including deformation, displacement, and flow of the slope material. Since our scene is represented by Gaussian primitives, the simulated particle motion can be naturally transferred to the corresponding Gaussian representation, enabling effective rendering of the slope dynamics at different time steps.

\section{Experiments}
\label{sec:experiments}

We evaluate the proposed UAV-based scan-to-simulation framework on a real slope scene and demonstrate its ability to produce physically plausible and visually realistic landslide dynamics from UAV observations. For physical simulation, we adopt the \textit{sand} material model from~\cite{xie2024physgaussian} within the MPM solver. The external force is set to standard gravitational acceleration, i.e., \(9.8~\mathrm{m/s^2}\). The material parameters are configured as follows: the friction angle is set to \(22^\circ\), the density to \(2000~\mathrm{kg/m^3}\), the Young's modulus to \(5\times10^7~\mathrm{Pa}\), and the Poisson's ratio to \(0.3\). We define a cuboid simulation domain that encloses the target slope region for analysis. For the boundary surfaces, we impose slip boundary conditions, allowing particles to slide along the surfaces instead of being fully fixed, which is suitable for modeling downhill soil movement. 


The entire framework is executed on a machine equipped with an NVIDIA RTX 4090 GPU. As can be observed in Figure~\ref{fig:result}, our method successfully simulates landslide behavior in a specified local region of the slope. At the same time, because the simulation is built on top of the Gaussian scene representation, the rendered results preserve the visual realism of the original reconstructed slope, including geometry-aware appearance and coherent scene context. Overall, the results demonstrate that our framework can effectively connect UAV-based slope sensing, Gaussian scene reconstruction, and physics-informed landslide simulation into a unified pipeline, highlighting the potential of physical AI for civil infrastructure scenarios where digital scene capture and predictive physical modeling need to work together for risk assessment and analysis.
\section{Conclusion}
\label{sec:conclusion}

In this paper, we presented a UAV-based scan-to-simulation framework for landslides using physics-informed Gaussian splatting. The proposed framework demonstrates the potential of bridging neural scene representations and physics-based simulation for civil infrastructure applications. In the future, this work can be extended in several directions. First, it would be valuable to incorporate more advanced and domain-specific geomechanical constitutive models to better capture the complex behavior of slope materials. Second, further automation of the overall pipeline would improve usability and scalability in practical deployment. Third, supporting simulation regions with more diverse geometric extents, beyond simple manually defined volumes, could enable more flexible and accurate landslide analysis in complex terrains.
{
    \small
    \bibliographystyle{ieeenat_fullname}
    \bibliography{main}
}


\end{document}